\documentclass{article}

\usepackage[nonatbib]{neurips_2020}

\usepackage[utf8]{inputenc} 
\usepackage[T1]{fontenc}    
\usepackage{hyperref}       
\usepackage{url}            
\usepackage{booktabs}       
\usepackage{amsfonts}       
\usepackage{nicefrac}       
\usepackage{microtype}      

\usepackage{graphicx}
\usepackage{amsmath}

\usepackage{geometry}
\graphicspath{ {./images/} }
 
\title{Robust Reinforcement Learning on Graphs for Logistics optimization}

\author{
  Zangir Iklassov \\
  MBZUAI \\
  PhD in Machine Learning \\
  20020082@mbzuai.ac.ae \\
  \And
  Dmitrii Medvedev \\
  MBZUAI \\
  MSc in Machine Learning \\
  20020035@mbzuai.ac.ae \\
}

\begin{document}

\maketitle

\begin{abstract}

Logistics optimization nowadays is becoming one of the hottest  areas in the AI community. In the past year, significant advancements in the domain were achieved by representing the problem in a form of graph. Another promising area of research was to apply reinforcement learning algorithms to the above task. In our work, we made advantage of using both approaches and apply reinforcement learning on a graph. To do that, we have analyzed the most recent results in both fields and selected SOTA algorithms both from graph neural networks and reinforcement learning. Then, we combined selected models on the problem of AMOD systems optimization for the transportation network of New York city. Our team compared three algorithms - GAT, Pro-CNN and PTDNet - to bring to the fore the important nodes on a graph representation. 
Finally, we achieved SOTA results on AMOD systems optimization problem employing PTDNet with GNN and training them in reinforcement fashion.

\end{abstract}

\section{Introduction}

For a long time, the problem of logistics optimization has used classical brute force algorithms or MPC. These algorithms are inefficient in time, but accurate. According to the research of McKinsey \cite{mckinsey}, logistics is one of the industrial areas in which machine learning is expected to make one of the greatest progresses and can save huge costs in the next decade. For this, machine learning algorithms shall become more accurate than classical algorithms, while being already more time-efficient, since they solve the problem in polynomial time, in contrast to exponential time of classical algorithms. One of the potential directions in this field is reinforcement learning. It has already shown striking results for strategic tasks such as chess, Go, Atari \cite{atari} \cite{alphazero}.

Since logistics is also a strategic optimization problem, the potential for using reinforcement learning for it is quite promising. To do this, RL should show robust results on different settings, on data of different types (homogeneous and heterogeneous) and different sizes (city, state, global scale). We concentrate on the second problem, where it is necessary to develop an algorithm which can build a logistic plan of a high quality independently of a given scale. Since the problem can be perfectly represented in the form of a graph, this shall also allow using the same model on the data of different scale, we are aiming to use a combination of RL and GNN to verify the robustness of the model on logistics (AMOD) problem. For now, our team has already analyzed the existing literature on this topic, reimplemented SOTA models on different datasets and checked their robustness on different graph sizes of New York transport environment.

We are planning to use such SOTA models as GAT, Pro-GNN and PTDNet to improve data size robustness of RL on AMOD problem. To the best of our knowledge, there was no research conducted before in this direction.

\section{Background}

Reinforcement learning is one of the methods of Machine Learning in which the model does not have any information about the system but has an option to interact with it and learn the policy from this experience. To formalize the reinforcement learning process we shall refer to Markov Decision Process (MDP) $M$ which can be described as a function of a space $S$ of system's observed states $s \in S$, actions $a \in A$ available to the agent in that space, conditional probability $P(s_{t+1} | s_t, a_t )$ of being in state $s_{t+1}$ upon taking action $a_t$ at the state $s_t$, initial distribution $d_0$ (starting point), reward $r$ to identify the goodness of the steps which agent takes, and a discount factor $0 < \gamma \leq 1$:

$$ M = (S, A, P, d_0, r, \gamma) $$

The  ultimate goal of reinforcement learning is to derive the distribution of optimal actions to take in each state, also called $\textit{policy}$ $\pi(a_t | s_t) $. In order to do that, the agent interacts with MDP following initial policy which is to be adjusted by the cycle: observing states $s_t$, taking actions $a_t$ at that state and being rewarded or punished by $r_t$ depending on the newly observed state $s_{t+1}$.

Graph Neural Networks (GNNs)are special type of neural network which directly works with a graph structure. Graph is a data structure $G$ which consists of a set of vertices (nodes) $V$ and edges $E$ connecting them:

$$ G = (V, E) $$

It is often very convenient to represent one's data as a graph structure when performing an image analysis task. This conveniences is directly related to graph's property of permutation invariance, meaning that an order of vertices is not affecting the output when performing calculations on them. Permutation invariance dramatically decreases the number of training examples required to generalize the model. However for traditional machine learning techniques one first need to represent graph-structured data into numerical vectors or other types of data structures, and such representation may lead to partial loss of data. GNN allows to work with graph data directly avoiding the above mentioned inconvenience. The core idea of GNN is based on a propagation of information on a graph. The latter is processed by a set of modules which are interconnected in accordance with the graph's edges and also linked with the vertices. While training, these modules update their states and exchange information until they reach an equilibrium. The mechanism to propagate information is limited to make sure the state of equilibrium exists. The output of GNN is calculated based on the state of the module at each vertex.

On of the types of GNN is Graph Convolutional Network (GCNs) used for image classification. GCNs use the following function to propagate information on a graph:

$$ X'= f(X, A) = \sigma(\hat{D}^{-\frac{1}{2}}\hat{A}\hat{D}^{-\frac{1}{2}}XW) \ ,\ \hat{A} = A + I $$

Where $I$ is an identity matrix, $\hat{D}$ is a diagonal node degree matrix of $\hat{A}$, $\sigma$ is a non-linear activation function and $W$ is a matrix of weights.

\section{Literature Review}

In the field of logistics optimization, the reinforcement learning approach is steadily gaining popularity among other machine learning techniques for its promising results. In 2017 Jian Wen et al. in \cite{jianwen} tackled rebalancing needs of AMOD systems. The team employed RL to be able to manage on large systems where data might be not fully available. As the result, the computational speed was 2.5 times faster versus classical models used in logistics while the performance of the model demonstrated near-optimal solution behavior. As the result, RL applications was considered a success for the task providing benefits for both users and operators.

In \cite{gnn_amod} Daniele Gammelli et al. presented the first work that combined reinforcement learning and graph neural network on AMOD systems. They represented the transportation network as graph with areas of the city as vertices and connectivity between those areas as edges. RL algorithm was applied on a graph to manage the rebalancing of AMOD systems. The team showed that GNN provides the basis for reinforcement learning agent to outperform classical models in transferability, generalization and ability to scale when solving logistics optimization task. Some of the achieved results of the work are shown in Table 1. We can see that the algorithm proposed in \cite{gnn_amod} demonstrated close-to-optimal performance and generate more than 36\% cost savings versus learning-based approaches of algorithms with classical architectures.

\begin{table}[h!]
  \caption{SYSTEM PERFORMANCE ON NEW YORK 4 × 4 NETWORK}
  \label{table1}
  \centering
  \begin{tabular}{lccc}
    Model     & Reward  & Demand    & Rebalancing \\
             & (\%Dev. MPC-tri-level) & Served    & Cost (\$) \\
    \midrule
    ED &	30,746 (-10.7\%) & 8,770 & 7,990     \\
    CQL &	30,496 (-11.4\%) & 8,736 & 8,284     \\
    A2C-MLP & 30,664 (-10.9\%) & 8,773 & 7,920     \\
    A2C-CNN & 30,443 (-11.5\%) & 8,904 & 8,775    \\
    A2C-GNN (\textbf{Daniele Gammelli et al.}) & 33,886 (-1.6\%) & 8,772 & 5,038     \\
    MPC-tri-level & 34,416 (0\%) & 8,865 & 4,647     \\
    MPC-standard & 35,356 (2.7\%) & 8,968 & 4,296    \\
    \midrule
    A2C-GNN-0Shot & 33,397 (-3.0\%) & 8,628 & 4,743     \\
    \bottomrule
  \end{tabular}
\end{table}

The most recent work in the field of reinforcement learning for logistics solutions is done by Yimo et al. in \cite{yimoyan}. The paper provides comprehensive overview of RL methods and classifies previous studies in the field. Among the strengths, the team highlights the ability of RL to learn from historical data for prediction purposes, as well as provide forecasting and optimisation approaches for stochastic tasks in logistics. At the same time reinforcement learning faces issues when dealing with complex multi-agent systems, however this area of logistics does not concern MOD, rather facilities settings etc.

\section{Methodology}

For sparse graphs A2C-GNN shows SOTA results for Autonomous Mobility-on-Demand problem. Deviation from MPC solution tend to reach zero value for small grid dimensions \cite{gnn_amod}. This means that for a graph with small number of connections for each node, A2C-GNN reaches the ideal NP-hard solution faster than other algorithms. However, the same study shows that as grid dimensions increase, the difference with the ideal solution grows. That is, the less sparse the input graph, the worse the results of the A2C-GNN algorithm we have. This trend will not allow the algorithm to be used in the industry for large scale problems. It is important to make an algorithm robust to increase in graph complexity.

The ideal solution to the AMOD problem is a set of vehicles and assigned routes which is NP-hard task that can be ideally solved by MPC-tri-level algorithm. For each vehicle-node pair we choose only one node at a time as a partial route, while other nodes are meaningless at this iteration. However, when we use GNN algorithm we average information from all connected nodes. And the more connected nodes we have, the less meaningful becomes an information from every separate node, which makes it more difficult to find an ideal solution. We assume this is the reason why we are moving away from the ideal solution when we have more complex input graph. And to solve the problem, we should focus only on the most important nodes and consider only information coming from them, making averaged embeddings more meaningful for solving an NP-hard problem.

To do this, we will consider three methods that allow us to emphasize the importance of certain nodes and zero out the information of others. The first algorithm is well-known Graph attention network \cite{gat}, which finds contextual weights for each node, thus changing the contribution of each node to the calculation of the new embedding. The second algorithm is Pro-GNN \cite{pro} which optimizes the adjacency matrix making it more sparse and low rank for more robust GNN computations. The third algorithm is PTDNet \cite{ptd} which uses two neural networks, one to select the sparser subgraph and the other one, GNN, to process the subgraph making the final prediction.

\subsection{GAT}

GNN recomputes feature vector of each node $h$ given the formula: 

$$
\vec{h}_{i}^{\prime}=\sigma\left(\frac{1}{K} \sum_{k=1}^{K} \sum_{j \in \mathcal{N}_{i}} \alpha_{i j}^{k} \mathbf{W}^{k} \vec{h}_{j}\right)
$$

Where $a$ is an attention value which measures ‘importance’ of each adjacent node for a new feature vector, and can be expressed using the following three formulas:

$$
e_{i j}=a\left(\mathbf{W} \vec{h}_{i}, \mathbf{W} \vec{h}_{j}\right)
$$

$$
\alpha_{i j}=\operatorname{softmax}_{j}\left(e_{i j}\right)=\frac{\exp \left(e_{i j}\right)}{\sum_{k \in \mathcal{N}_{i}} \exp \left(e_{i k}\right)}
$$

$$
\alpha_{i j}=\frac{\exp \left(\operatorname{LeakyReLU}\left(\overrightarrow{\mathbf{a}}^{T}\left[\mathbf{W} \vec{h}_{i} \| \mathbf{W} \vec{h}_{j}\right]\right)\right)}{\sum_{k \in \mathcal{N}_{i}} \exp \left(\operatorname{LeakyReLU}\left(\overrightarrow{\mathbf{a}}^{T}\left[\mathbf{W} \vec{h}_{i} \| \mathbf{W} \vec{h}_{k}\right]\right)\right)}
$$

Here $\mathrm{W}$ is a weight matrix to be trained and both $i$ and $j$ represent indices of two different nodes. Attention weight $a$ is in range [0, 1] so the higher it is, the more important corresponding adjacent node becomes.

\subsection{Pro-GNN}

$$
\min _{\theta} \mathcal{L}_{G N N}\left(\theta, \mathrm{A}, \mathrm{X}, y_{L}\right)=\sum_{v_{i} \in \mathcal{V}_{L}} \ell\left(f_{\theta}(\mathrm{X}, \mathrm{A})_{i}, y_{i}\right)
$$

By default, we optimize the above function, where function $f$ is an output of GNN model with adjacency matrix $\mathrm{A}$, feature input $\mathrm{X}$ and weight matrix $\mathrm{W}$ to be trained.

$$
f_{\theta}(\mathbf{X}, \mathbf{A})=\operatorname{softmax}\left(\hat{\mathrm{A}} \sigma\left(\hat{\mathrm{A}} \mathbf{X} \mathbf{W}_{1}\right) \mathbf{W}_{2}\right)
$$

From now on we assume that $\mathrm{A}$ is over complicated, it is having inflated rank for efficient GNN use, and we shall replace it with another matrix $\mathrm{S}$ which we get from:

$$
\underset{\mathrm{S} \in \mathcal{S}}{\arg \min } \mathcal{L}_{0}=\|\mathrm{A}-\mathrm{S}\|_{F}^{2}+\alpha\|\mathrm{S}\|_{1}+\beta\|\mathrm{S}\|_{*}, \text { s.t., } \mathrm{S}=\mathrm{S}^{\top}
$$

Here we initialize $\mathrm{S} = \mathrm{A}$, then we optimize $\mathrm{S}$ by making it sparser and downgrade the rank by minimizing its $l_1$ and nuclear norms, while preserving it being symmetric and keeping it close to initial $\mathrm{A}$ as much as possible.

\subsection{PTDNet}

Finding output $Y$, given graph $G$, can be formulated in terms of several subgraphs $g$:

$$
P(Y \mid G) \approx \sum_{g \in \mathbb{S}_{G}} P(Y \mid g) P(g \mid G)
$$

Then, we can approximate second part using two different neural networks:

$$
\sum_{g \in \mathbb{S}_{G}} P(Y \mid g) P(g \mid G) \approx \sum_{g \in \mathbb{S}_{G}} Q_{\theta}(Y \mid g) Q_{\phi}(g \mid G)
$$

First network aims to select certain edges from initial adjacency matrix $\mathrm{A}$ forming subgraph $g$. Second network, given subgraph $g$, predicts output $Y$. Two networks can be trained together, given $G$ and $Y$; the only thing to consider is to make first part differentiable. For this purpose we will use Gumbel-Softmax function \cite{gumbel} that helps us to generate differentiable samples from a set of all the edges.

All three algorithms choose certain nodes over others emphasizing the importance of the formers, which can lead to more representative feature vectors of the nodes in terms of finding optimal routes and combinations for AMOD problem. It is possible to combine the above methods and study whether we can sum-up their effects, and this is to be one of the plans for our future research. However, in this work we will try firstly to check  separate effects of the those methods in order to understand the contribution of each algorithm to the AMOD problem.

\section{Results}

The environment used for both training and testing is the Manhattan district represented in k x k grids map, where each grid is one of the New York City stations. Firstly, we create a graph of these grids and connect adjacent stations. Secondly, we run a simulation in which a daily demand for commute is synthesized. The model has a number of vehicles at its disposal which it has to assign to meet a demand and assign routes in order to maximize daily income. We can represent Manhattan in a different number of stations as a k x k grid. The larger k we have, the more complex graph we will build.

On the input, the model receives a graph in which nodes are represented as stations and edges are connections between those stations. Each edge stores the information about the cost of moving from one node to another and the number of passengers who want to move this path. Each node stores information about the number of vehicles available in that station. The matrix of edges $\mathrm{A}$ and matrix of features $\mathrm{X}$ are fed to the Graph neural network that has one layer of graph convolution with ReLU activation function followed by 3 fully-connected layers with 32 neurons each. Actor and Critic networks have the same architecture with the only difference in output layer. Actor produces actions represented as a percentage distribution of vehicles to be rebalanced for each node at a given time step, that is predicting value in the range [0, 1] for each station. Critic aggregates information from all nodes using sum-pool function and predicts one value for the whole graph, where reward is represented as a profit at each time step.

For GAT network we replaced GCN layer with Graph Attention layer, for Pro-GNN network we added optimization of adjacency matrix described in Section 4.2. Finally, for PTDNet we built additional sampling network that consists of several MLP layers and for every edge predicts probability of leaving it in A, changing A that will be used then in GCN.

To train the models we employed PyTorch library in Python and built RL model. The model was trained using the Adam optimizer with 0.003 learning rate and 0.97 discount factor, for 16,000 episodes. To evaluate performance, we used demand, expressing it in total amount of executed trips, cost, expressing it in the expenses for all movements, and deviation from MPC-tri-level, which is an ideal solution for this problem, but unusable for a large real-life values of k.

\begin{table}[h!]
  \caption{Performance on New York 4 × 4 Network}
  \label{table2}
  \centering
  \begin{tabular}{lccc}
    Model     & Reward  & Demand    & Rebalancing \\
             & (\%Dev. MPC-standard) & Served    & Cost (\$) \\
    \midrule
    A2C-GNN &	33,384 (-3\%) & 8,699 & 5,036     \\
    A2C-GAT &	33,280 (-3,3\%) & 8,569 & 5,049     \\
    A2C-Pro-GNN &	33,246 (-3,4\%) & 8,664 & 5,018     \\
    A2C-PTDNet &	33,418 (-2,9\%) & 8,709 & 5,031     \\
    \bottomrule
  \end{tabular}
\end{table}


\begin{table}[h!]
  \caption{Performance on New York 20 × 20 Network}
  \label{table3}
  \centering
  \begin{tabular}{lccc}
    Model     & Reward  & Demand    & Rebalancing \\
             & (\%Dev. MPC-standard) & Served    & Cost (\$) \\
    \midrule
    A2C-GNN &	131,701 (39) & 39,825 & 28,151    \\
    A2C-GAT &	130,288 (37.6) & 39,235 & 28,219    \\
    A2C-Pro-GNN &	134,152 (31.7) & 41,694 & 27,324     \\
    A2C-PTDNet &	138,8365 (30.1) & 42,291 & 27,019     \\
    \bottomrule
  \end{tabular}
\end{table}

The tables 1, 2 show the performance of all models for 4x4 and 20x20 grids' environments. In the first case, the differences between the models are insignificant, all performance metrices vary around the same values. However, in the second case, the difference rose sharply, with all three new models showing a better result than the baseline and the best result belongs to A2C-PTDNet with the smallest deviation of 30.1\%, the largest demand and the smallest cost of 42,291 and 27,019 values respectively. The larger the value of k we choose, the higher the difference between the models’ performance we have. It can be seen in Figure 1.

\begin{figure}[h!] 
    \centering
    \includegraphics[width=8cm]{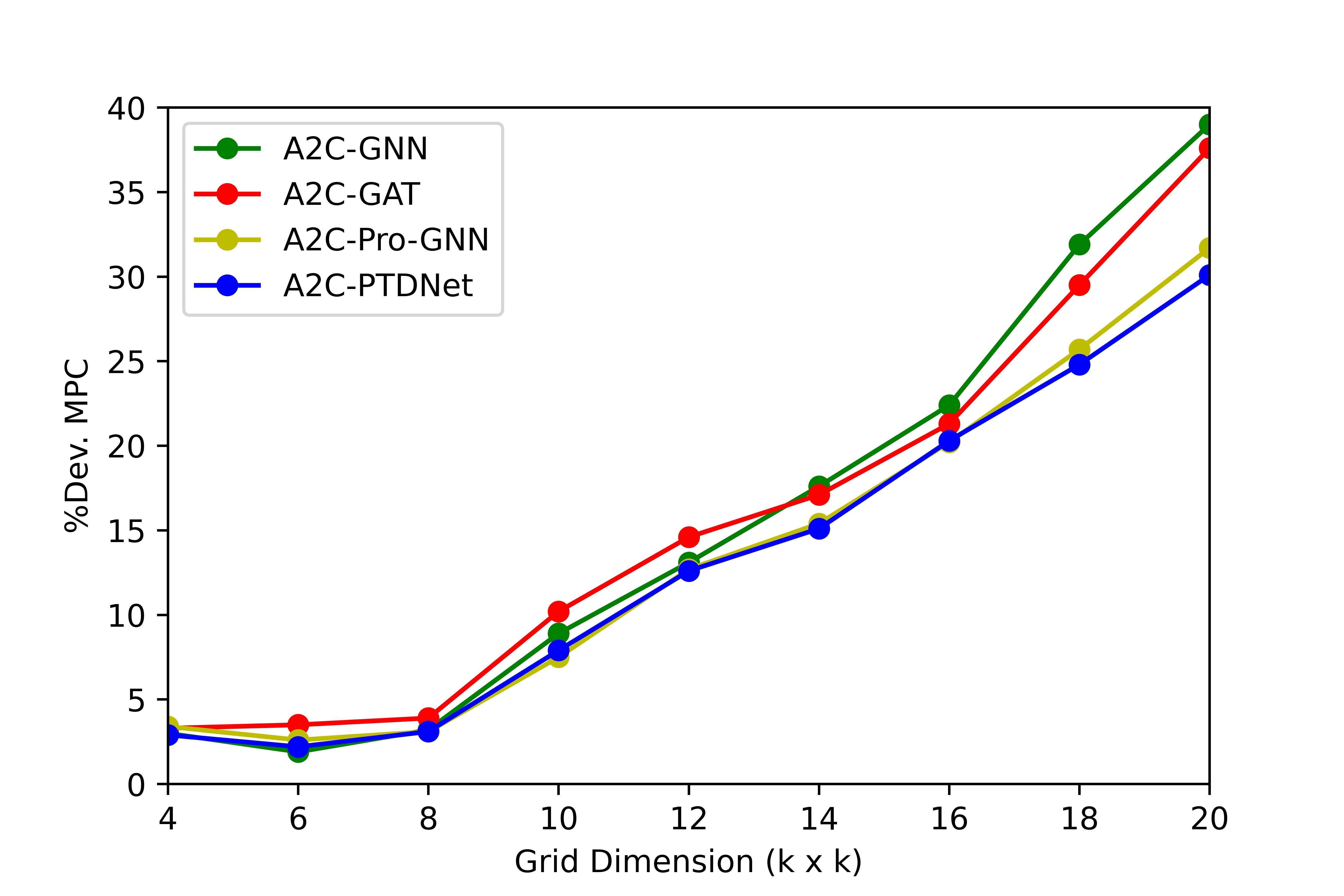}
    \caption{Performance of models trained on single granularity (4 × 4).}
\end{figure}

\section{Discussion}

The results show an improvement in model’s performance. However, the reason for this may be that the model, in all three cases, has become more complex in comparison with the base model as a whole and not more robust to the size of the setting in particular. In addition to the above, the negative relationship between the quality of prediction and the size of the setting still remained quite high. For the final solution of this problem, this dependence should be close to zero. However, we have shown that focusing on ‘most important’ nodes during calculations can improve the performance. Moreover, discrete selection of nodes works better than their weighting or adjacency matrix continuous transformation. Thus, indeed, an increase in the average number of connected nodes for large graphs negatively affects learning, due to the greater blurring of information received from each node. In the future research, it can be effective to focus on a discrete selection of nodes for network performance improvement, as well as to test other techniques of node selection and nodes combinations, and to test the models on other logistics tasks besides AMOD.

\section{Conclusion}

We got new SOTA results on the New York dataset of AMOD problem using PTDNet with GNN architecture trained in reinforcement fashion. However, it is necessary to further improve the result in order to obtain a robust result no matter of setting size and the closest possible solution to MPC-tri-level (ideal). In this case, the model will be beneficial for use in the industry, and can become in demand and save large costs for logistics tasks. In general, this direction is potentially beneficial and interesting from both theoretical and practical points of view.






\newpage
\begin{thebibliography}{8}

\bibitem{gnn_amod}
Gammelli, D., Yang, K., Harrison, J., Rodrigues, F., Pereira, F. C., \& Pavone, M. (2021). Graph Neural Network Reinforcement Learning for Autonomous Mobility-on-Demand Systems. arXiv preprint arXiv:2104.11434.

\bibitem{gat}
Veličković, P., Cucurull, G., Casanova, A., Romero, A., Lio, P., \& Bengio, Y. (2017). Graph attention networks. arXiv preprint arXiv:1710.10903.

\bibitem{pro}
Jin, W., Ma, Y., Liu, X., Tang, X., Wang, S., \& Tang, J. (2020, August). Graph structure learning for robust graph neural networks. In Proceedings of the 26th ACM SIGKDD International Conference on Knowledge Discovery \& Data Mining (pp. 66-74).

\bibitem{ptd}
Luo, D., Cheng, W., Yu, W., Zong, B., Ni, J., Chen, H., \& Zhang, X. (2021, March). Learning to drop: Robust graph neural network via topological denoising. In Proceedings of the 14th ACM International Conference on Web Search and Data Mining (pp. 779-787).

\bibitem{gumbel}
Jang, E., Gu, S., \& Poole, B. (2016). Categorical reparameterization with gumbel-softmax. arXiv preprint arXiv:1611.01144.

\bibitem{atari}
Mnih, V., Kavukcuoglu, K., Silver, D., Graves, A., Antonoglou, I., Wierstra, D., \& Riedmiller, M. (2013). Playing atari with deep reinforcement learning. arXiv preprint arXiv:1312.5602.

\bibitem{alphazero}
Silver, D., Hubert, T., Schrittwieser, J., Antonoglou, I., Lai, M., Guez, A., ... \& Hassabis, D. (2018). A general reinforcement learning algorithm that masters chess, shogi, and Go through self-play. Science, 362(6419), 1140-1144.

\bibitem{mckinsey}
Baer, T., \& Kamalnath, V. (2017). Controlling machine-learning algorithms and their biases. McKinsey Insights.

\bibitem{jianwen}
Wen, J., Zhao, J., \& Jaillet, P. (2017, October). Rebalancing shared mobility-on-demand systems: A reinforcement learning approach. In 2017 IEEE 20th International Conference on Intelligent Transportation Systems (ITSC) (pp. 220-225). Ieee.

\bibitem{yimoyan}
Yan, Y., Chow, A. H., Ho, C. P., Kuo, Y. H., Wu, Q., \& Ying, C. (2021). Reinforcement Learning for Logistics and Supply Chain Management: Methodologies, State of the Art, and Future Opportunities. State of the Art, and Future Opportunities (October 4, 2021).

\end{thebibliography}
\end{document}